# SEMI-SUPERVISED DEEP LEARNING FOR HIGH-DIMENSIONAL UNCERTAINTY QUANTIFICATION

**Zequn Wang[1] and Mingyang Li[2]**

Michigan Technological University, Houghton, MI, USA

**ABSTRACT**

*Conventional uncertainty quantification methods usually lacks the capability of dealing with high-dimensional problems due to the curse of dimensionality. This paper presents a semi-supervised learning framework for dimension reduction and reliability analysis. An autoencoder is first adopted for mapping the high-dimensional space into a low-dimensional latent space, which contains a distinguishable failure surface. Then a deep feedforward neural network (DFN) is utilized to learn the mapping relationship and reconstruct the latent space, while the Gaussian process (GP) modeling technique is used to build the surrogate model of the transformed limit state function. During the training process of the DFN, the discrepancy between the actual and reconstructed latent space is minimized through semi-supervised learning for ensuring the accuracy. Both labeled and unlabeled samples are utilized for defining the loss function of the DFN. Evolutionary algorithm is adopted to train the DFN, then the Monte Carlo simulation method is used for uncertainty quantification and reliability analysis based on the proposed framework. The effectiveness is demonstrated through a mathematical example.*

**INTRODUCTION**

Uncertainty is inherent to real-world engineering systems, and the task of rigorously quantifying the effect of input parameter uncertainty on the system responses is known as the uncertainty quantification or propagation. In engineering applications, reliability is defined as the probability that a system can perform its intended functionality with the consideration of uncertainties. Therefore, reliability analysis is of critical importance in the development of engineering systems as it quantifies the uncertainties such as the randomness of material properties, geometry, and environmental conditions.

The first- and second-order reliability methods (FORM and SORM) [1-3] are known as the typical analytical-based reliability analysis methods, which have been extensively studied for the past decades. The limit state functions are approximated through Taylor expansions, and reliability is estimated by finding the most probable point in a standard normal space. Due to the lack of accurate sensitivity information, non-convergence issue may occur. Thus, sampling-based methods such as Monte Carlo simulation (MCS) have been developed for improving the accuracy of reliability estimation. To alleviate the computational costs due to the large number of system responses evaluations, easy-to-evaluate surrogate models have been utilized as substitutes for computationally expensive simulations or experiments. Popular choices for surrogate models in the literature include, support vector machines (SVM) [4-7], Kriging models [8-10], and artificial neural networks [11-14]. Given a set of training data, surrogate models can be constructed and then MCS can be directly carried out for reliability analysis. Research efforts have been devoted to developing adaptive sampling strategies [15-18], which aim at balancing the fidelity of the surrogate model and the costs of function evaluations. For adaptive Kriging-based methods, an initial surrogate model of the limit state function is first constructed based on an initial set of training data, then it is sequentially updated by adding critical training samples. Wang and Wang [19] developed a maximum confidence enhancement method to iteratively search for most useful samples that can maximize the accuracy improvement of the Kriging model. Echard et al. [20] and Zhao et al. [21] considered the Kriging prediction variance in their sampling criterion for selecting additional training data. Dubourg et al. [22] developed an adaptive refinement technique to reduce the prediction errors, where the Kriging model can be updated by simultaneously adding multiple samples. Despite the success of surrogate models and adaptive sampling strategies, most of existing methods become intractable for high-dimensional problems.

To alleviate the curse of input dimensionality, various methods [23, 24] such as Karhunen-Loeve expansion [25], t-Distributed Stochastic Neighbor Embedding (t-SNE) [26], and high-dimensional model representation (HDMR) methods [27-29] have been applied for dimension reduction. Specifically, the HDMR methods are utilized to improve the performance of reliability analysis for high-dimensional problems, which aim at decomposing a high-dimensional performance function into multiple low-dimensional component functions. The low-dimensional functions are then approximated through interpolation techniques such as Gaussian quadrature. Based on the summation of the approximated low-dimensional functions, a global surrogate model of the performance function can be constructed accordingly. As a result, the number of function evaluations can be significantly reduced. Acar et al. [30]

---
[1] Assistant Professor, zequnw@mtu.edu, Corresponding Author
[2] Graduate Student, mli7@mtu.edu

employed the univariate dimension reduction method to decompose the multi-dimensional performance function, and then the estimated statistical moments are used to fit the parameters of extended generalized lambda distribution for calculating the reliability. Li et al. [31] integrated the expected improvement sampling strategy with dimension reduction method for solving complicated engineering problems. By increasing the order of component functions, the accuracy of reliability assessment can be improved, however, the number of function evaluations may increase dramatically. Though the HDMR is a feasible method for high dimensional problems, how to balance the prediction accuracy and the computational costs still remains a challenge. Moreover, most dimensionality reduction models are developed for unsupervised learning problems, which do not account for the regression task [32].

In the field of high-dimensional data analysis, deep learning [33-35] have gained lots of attention due to its capability of extracting critical features from high-dimensional space. Deep learning such as deep neural network and convolutional neural network are known as machine learning methods for learning data representations, and these techniques have been successfully applied in the fields of image processing [36], natural language understanding [37], robotic control [38], and high-energy physics [39]. In deep neural networks, a multivariate function can be modeled using a hierarch of features, where a series of nonlinear projections of the input is used to tackle the curse of dimensionality. Deep learning techniques can be classified into supervised, semi-supervise, and unsupervised learning according to the restriction on the training data. For instance, supervised learning requires labeled training data and unsupervised learning relies on unlabeled training data. Semi-supervised learning is an approach that combines both labeled and unlabeled data during the training process. Unsupervised learning is not applicable for uncertainty quantification due to the lack of system responses information. A large number of labeled data is usually required to properly train a deep neural network in supervised learning, which may result in unaffordable costs in handling practical engineering problems. In semi-supervised learning, the combination of both labeled and unlabeled data can lead to a considerable improvement in terms of learning accuracy. However, for the task of uncertainty quantification, how to systematically use both data sets in deep learning still remains a challenge.

This paper presents a semi-supervised learning method that employs evolutionary algorithm in order to deal with uncertainty quantification problems with high-dimensionality. The framework consists of three critical modules. First, an autoencoder neural network is trained based on a fused data set to introduce a low-dimensional latent space with distinguishable limit state, which is treated as a representative of the original high-dimensional space. To model the relationship between the low-dimensional representative and the system response, the second module, a surrogate model is established through Gaussian process regression. Thus, the high-dimensional uncertainty quantification is converted to a low-dimensional case. A deep feedforward neural network (DFN) is utilized to map high-dimensional system inputs to the low-dimensional space, where unlabeled data is provided to the three modules in a sequential manner, and eventually combined with the labeled data for updating the parameters of the DFN, lead to an enhancement of the learning accuracy. Monte Carlo simulation is then employed for reliability analysis in the low-dimensional space.

1. **Semi Supervised Learning for High-Dimensional Reliability Analysis**

This work aims at the uncertainty quantification for high-dimensional problems. Assuming the performance of an engineering system is modeled by a limit state function (LSF) $G(\mathbf{x})$, where $\mathbf{x} \in R^{nr}$ denotes the high-dimensional system input, and each input variable is considered as an independent random variable. Due to the randomness of the high-dimensional inputs, the proposed framework is utilized to estimate the stochastic system responses through semi-supervised learning. As shown in Fig.1, three critical modules are involved, including 1) an autoencoder to introduce a latent space for dimension reduction, 2) a GP model for linking the low-dimensional variables to the system responses, and 3) a deep feedforward neural network that is used for latent space reconstruction.

Based on a set of labeled data, the high-dimensional system input and their corresponding system responses are fused to train an autoencoder neural network, leading to a low-dimensional abstraction of the original high-dimensional space, referred to as latent space. With the achieved latent variables and the responses information, a GP model is built which is treated as the surrogate model that learns the relationship between the latent variables and the system responses. Thus, uncertainty quantification for the system responses can be conducted in the latent space instead of in the original high-dimensional space. To connect the high-dimensional inputs to the latent variables, a deep feedforward neural network is established without requiring extra information of system responses, where semi-supervised learning is performed to update the parameters of the DFN based on both the labeled and unlabeled data sets. The connection of the three critical modules can be summarized as

$$G(\mathbf{x}) = g(\mathbf{\theta}) = g(nn(\mathbf{x})) \qquad (1)$$

where $nn(.)$ represents the deep feedforward neural network, $\mathbf{\theta}$ represents the latent variables, $g(.)$ represents the GP surrogate model in the latent space. Though the autoencoder is not directly shown in Eq. (1), the mapping relationship provided by the autoencoder is the fundamental component for connecting the three critical modules. As a result, the three modules are utilized for uncertainty propagation and reliability analysis through Monte Carlo simulation. The structure of the three modules are introduced in subsection 1.1, and the details of the semi-supervised learning is provided in subsection 1.2. In subsection 1.3, the proposed framework is employed for high-dimensional reliability analysis.

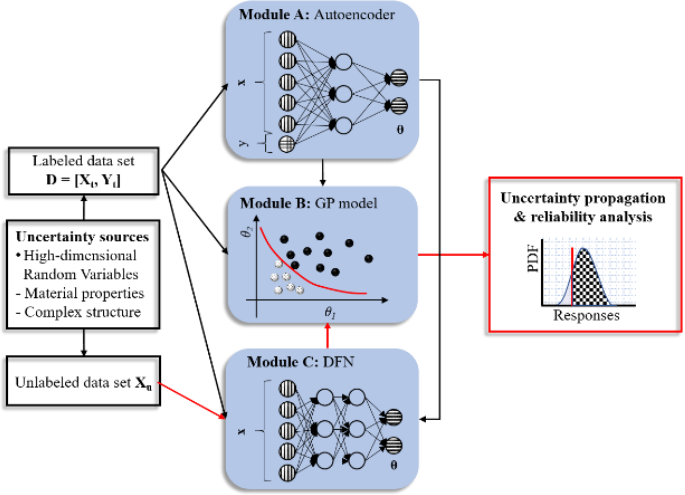

**FIGURE 1:** UNCERTAINTY QUANTIFICATION USING THE PROPOSED SEMI-SUPERVISED LEARNING

## 1.1 Critical Machine Learning Modules
### 1.1.1 Dimension reduction using autoencoder

An autoencoder neural network consists of an encoder and a decoder, where the encoder compresses the inputs to a low dimensional space and the decoder is then utilized to reconstruct the inputs. The low dimensional space is referred to as the latent space or latent layer, and the latent variables is used to denote the compressed representation of the inputs. In this paper, the purpose of utilizing the autoencoder mainly focuses on the encoder part, aims at mapping the high dimensional inputs to low dimensional latent variables.

Let $\mathbf{X_t} = [\mathbf{x}_1, \mathbf{x}_2, \ldots, \mathbf{x}_n]$ represents $n$ system input sites and $\mathbf{Y_t} = [y_1, y_2, \ldots, y_n]$ denotes the evaluated system responses at these input, the fused data set $\mathbf{D}$ is considered as the training input data for the autoencoder, which is expressed as

$$\mathbf{D} = \begin{bmatrix} x_1^{(1)} & \cdots & x_n^{(1)} \\ \vdots & \cdots & \vdots \\ x_1^{(nr)} & \cdots & x_n^{(nr)} \\ y_1 & \cdots & y_n \end{bmatrix} \qquad (2)$$

where $x_i^{(j)}$ denotes the $j^{th}$ input variable of the $i^{th}$ realizations of the system input, and $nr$ denotes the dimensionality of the input $\mathbf{x}$. According to Eq. (2), the fused data set $\mathbf{D}$ is a $n \times (nr + 1)$ matrix. In autoencoder, the training label is the same as the training input. Therefore, both the input and output dimension of the autoencoder are given as $(nr + 1)$. Assuming an autoencoder has $T$ total layers, where the $L^{th}$ layer is the latent layer ($L < T$), the computation of the $j^{th}$ layer in the encoder can be expressed as,

$$\mathbf{m}^{(j)} = f_{act-en}\left(\mathbf{W}_{en}^{(j)} \mathbf{m}^{(j-1)} + \mathbf{b}_{en}^{(j)}\right), \forall j \in \{1, 2, \ldots L\} \qquad (3)$$

where $\mathbf{m}^{(j)}$ represents the neurons in the $j^{th}$ hidden layer, $\mathbf{b}_{en}$ is a vector of bias and $\mathbf{W}_{en}$ represents the corresponding weight matrix, and $f_{act-en}$ is an activation function for the encoder. At the first layer ($j = 1$), the term $\mathbf{m}^{(1)}$ represents the training input $\mathbf{D}$, and $\mathbf{m}^{(L)}$ denotes the corresponding latent variables $\mathbf{\theta_t}$ in the latent space with dimensionality $nz$. In the decoder part, the goal is to reconstruct an input representation of the original input $\mathbf{D}$ based on the latent variables, and the calculation of the neurons has a similar form, written as

$$\mathbf{m}^{(j)} = f_{act-de}\left(\mathbf{W}_{de}^{(j)} \mathbf{m}^{(j-1)} + \mathbf{b}_{de}^{(j)}\right), \forall j \in \{L, \ldots T\} \qquad (4)$$

In the last layer, an input representation is achieved as $\mathbf{D'} = \mathbf{m}^{(T)}$. In Eq. (4), the term $f_{act-de}$ is an activation function used in the decoder part, and it can be either the same or different as the one utilized in the encoder part. The common choices for the activation function include the rectified linear unit (ReLU), logistic function, and hyperbolic tangent function (tanh). In this work, the logistic sigmoid function is adopted for $f_{act-en}$, which can be written as

$$f_{act}(\mathbf{m}) = \frac{1}{1+e^{-\mathbf{m}}} \qquad (5)$$

Once the layers and activation functions are defined for the autoencoder, the parameters such as weights and biases that fully characterize the network needs to be determined. Given the training data $\mathbf{D}$, these parameters can be estimated by minimizing the mean square error (MSE) between the original input and the reconstruction $\mathbf{D'}$, expressed as

$$\ell(\mathbf{D}, \mathbf{D'}) = \left\| \mathbf{D} - f_{act-de}\left(\mathbf{W}_{de} f_{act-en}\left(\mathbf{W}_{en}\mathbf{m} + \mathbf{b}_{en}\right) + \mathbf{b}_{de}\right) \right\|^2 \qquad (6)$$

Equation (6) is also known as the loss function. Once the autoencoder is trained, the corresponding latent variables $\mathbf{\theta_t}$ can be computed accordingly, which will be treated as a compressed representation of the fused high-dimensional data $\mathbf{D}$. Therefore, the encoder part of the autoencoder is utilized for dimension reduction, where the dimensionality of the problem is reduced from $(nr + 1)$ to $(nz)$. Moreover, the fused training data contains the information of both system inputs and responses, lead to a distinguishable limit state in the latent space.

### 1.1.2 GP modeling for LSFs in latent space

Gaussian process (GP) modeling is known as a typical nonparametric regression technique, and it has been widely applied in engineering applications due to its robustness and efficiency. In this paper, the GP modeling technique is employed to construct surrogate models for the limit state function in the latent space. For the GP model, the training data set is given as $\mathbf{\theta_t}$ and $\mathbf{Y_t}$, where $\mathbf{\theta_t} = [\mathbf{\theta}_1, \mathbf{\theta}_2, \ldots, \mathbf{\theta}_n]$ represents the actual latent variables that directly obtained from the autoencoder, and $\mathbf{Y_t}$ is the corresponding actual system responses that treated as the noisy observed output values. The noise is considered as a normally distributed random variable with zero men and standard deviation $\sigma_\varepsilon$. Therefore, the limit state function in the

[Type here]

latent space $g(\boldsymbol{\theta})$ can be modeled as a Gaussian process, which is expressed as

$$g(\boldsymbol{\theta}) \sim GP(m(\boldsymbol{\theta}), R(\boldsymbol{\theta}_i, \boldsymbol{\theta}_j)) \quad (7)$$

The covariance function $R(\boldsymbol{\theta}_i, \boldsymbol{\theta}_j)$ characterizes the correlation between the responses at points $\boldsymbol{\theta}_i$ and $\boldsymbol{\theta}_j$. A GP model is constructed once its mean and covariance functions are fixed. In this paper, the mean function is set to zero, and the squared exponential covariance function is adopted, which contains two hyperparameters and can be formulated as

$$R(\boldsymbol{\theta}_i, \boldsymbol{\theta}_j) = \alpha^2 \exp\left[-(\boldsymbol{\theta}_i - \boldsymbol{\theta}_j)^T \mathbf{P}^{-1}(\boldsymbol{\theta}_i - \boldsymbol{\theta}_j)/2\right] \quad (8)$$

where $\mathbf{P}$ is a diagonal matrix of an unknown parameter $\omega^2$, $\alpha$ represents the signal standard deviation, and $\omega$ represents the length-scale. With the consideration of the a Gaussian noise with zero men and standard deviation $\sigma_\varepsilon$, one can write the joint distribution of the training output and the response prediction at any input $\boldsymbol{\theta}^{e'}$ as

$$\begin{bmatrix} \mathbf{Y}_t \\ y' \end{bmatrix} \sim N\left(0, \begin{bmatrix} \mathbf{R}(\boldsymbol{\theta}_t, \boldsymbol{\theta}_t) + \sigma_\varepsilon^2 \mathbf{I} & \mathbf{R}(\boldsymbol{\theta}_t, \boldsymbol{\theta}^{e'}) \\ \mathbf{R}(\boldsymbol{\theta}^{e'}, \boldsymbol{\theta}_t) & \mathbf{R}(\boldsymbol{\theta}^{e'}, \boldsymbol{\theta}^{e'}) \end{bmatrix}\right) \quad (9)$$

where $\mathbf{R}(.,.)$ represents the covariance matrix obtained based on Eq. (8). To estimate the hyperparameter, the Gaussian likelihood function is adopted based on the training data set. As a result, the GP model is capable of predicting the system response given any estimated latent variable $\boldsymbol{\theta}^e$. The response prediction follows a normal distribution, where the prediction mean and variance are given as

$$\mu_{gp}(\boldsymbol{\theta}^e) = \mathbf{r}^T \left[\mathbf{R}(\boldsymbol{\theta}_t, \boldsymbol{\theta}_t) + \sigma_\varepsilon^2 \mathbf{I}\right]^{-1} \mathbf{Y}_t \quad (10)$$

and

$$v_{gp}(\boldsymbol{\theta}^e) = \mathbf{R}(\boldsymbol{\theta}^e, \boldsymbol{\theta}^e) - \mathbf{r}^T \left[\mathbf{R}(\boldsymbol{\theta}_t, \boldsymbol{\theta}_t) + \sigma_\varepsilon^2 \mathbf{I}\right]^{-1} \mathbf{R}(\boldsymbol{\theta}_t, \boldsymbol{\theta}^e) \quad (11)$$

where $\mathbf{r}$ is the correlation vector between the existing training points and the input point $\boldsymbol{\theta}^e$.

For the GP model, the training data set is given as $\boldsymbol{\theta}_t$ and $\mathbf{Y}_t$, where $\boldsymbol{\theta}_t = [\boldsymbol{\theta}_1, \boldsymbol{\theta}_2, \ldots, \boldsymbol{\theta}_n]$ represents the actual latent variables that directly obtained from the autoencoder, and $\mathbf{Y}_t$ is the corresponding actual system responses that treated as the noisy observed output values.

### 1.1.3 DFN for latent space reconstruction

With the autoencoder, one can always obtain latent variables by providing the combination of system input $\mathbf{x}$ and its corresponding response $y$. However, evaluating system responses are usually expensive due to the repeated runs of experiments and large-scale simulations. I nstead of directly evaluating the system responses for uncertainty quantification, we aims at reconstructing the latent space by providing only the input parameters $\mathbf{x}$. It is worth noting that the distribution of the latent variables in the low-dimensional space contains the information from both the system inputs and responses. Therefore, the goal of latent space reconstruction is to estimate the latent variables without knowing the actual system responses.

A feedforward neural network consists of three types of layers, including the input, hidden, and output layers, where the size of each layer is defined by the number of neurons. For instance, the number of neurons on the first and last layer is equal to the dimensionality of the training input and training labels, respectively. Each hidden layer contains a set of neurons that are completely independent of each other, where each neuron is fully connected to all neurons in the previous layer. The computation of a single neuron that connected to $p$ neurons in the previous layer is expressed as

$$m' = f_{act}\left(\sum_{i=1}^{p}(w_i m_i + b_i)\right) \quad (12)$$

where $w_i$ and $b_i$ represent the weight and bias of the $i^{th}$ neurons, respectively. As shown in Eq. (12), each neuron in the previous layer is multiplied by a weight $w_i$ then followed by bias, the summations are passed to an activation function $f_{act}(.)$. In general, the rectified linear unit (ReLU), hyperbolic tangent function (tanh), and logistic sigmoid function are common choices of the activation function. For supervised learning, the training data set including both the training inputs and labels needs to be provided to the DFN. Usually the mean square errors that capture the mismatch between the training labels and the predictions is adopted as the loss function during the training process. The loss function will be minimized through backpropagation algorithm to determine the weights and biases.

In this work, the DFN is served as a link function which learns the mapping relationship between the system input $\mathbf{x}$ and the corresponding latent variables, which can be expressed as

$$\boldsymbol{\theta}^e = nn(\mathbf{x}) \quad (13)$$

where $\boldsymbol{\theta}^e$ is used to represent the latent variables that estimated by the DFN, while $\boldsymbol{\theta}_t$ represents the latent variables directly obtained from the autoencoder. Instead of using supervised learning to train the DFN based on labeled data set $\mathbf{D}$, a semi-supervised learning method is proposed for enhancing the accuracy of uncertainty quantification, which will be introduced in the following subsection.

### 1.2 Semi-Supervised Learning Using Evolutionary Algorithm

To ensure the accuracy of the latent space reconstruction without incurring extra computational costs, a semi-supervised learning procedure is introduced in this work. The core idea lies in that an additional set of unlabeled samples are merged with the labeled data set to iteratively update the deep feedforward neural network. As shown in Fig. 1, the input layer for the DFN is the system input $\mathbf{x}$, and the output layer is the latent variables $\boldsymbol{\theta}$. Therefore, the inputs and outputs of the labeled data set are $\mathbf{X}_t = [\mathbf{x}_1, \mathbf{x}_2, \ldots, \mathbf{x}_n]$ and $\boldsymbol{\theta}_t = [\boldsymbol{\theta}_1, \boldsymbol{\theta}_2, \ldots, \boldsymbol{\theta}_n]$, respectively. Let $\mathbf{X}_u = [\mathbf{x}_1, \mathbf{x}_2, \ldots, \mathbf{x}_q]$ denote $q$ samples that are generated according to the randomness of the system input variables $\mathbf{X}$. The corresponding system responses $\mathbf{Y}_u = G(\mathbf{X}_u)$ are unknown. Without the corresponding system responses information, the trained autoencoder cannot be applied for computing the latent variables $\boldsymbol{\theta}_u$ that is corresponding to $\mathbf{X}_u$. Therefore, samples $\mathbf{X}_u$ are referred to as unlabeled samples.

The procedure of the proposed semi-supervised learning is shown in Fig. 2, where different types of arrows are used to indicate the information flow for labeled and unlabeled data, respectively. To deal with the unlabeled data, the deep feedforward network can be utilized for estimating these latent variables given only the system inputs, expressed as

$$\boldsymbol{\theta}_u^e = nn(\mathbf{X}_u) \quad (14)$$

Then the GP model $g(.)$ that is built in the latent space can be employed for predicting the system responses based on the estimated latent variables. As a result, the system responses of the unlabeled data can be approximated through the latent space reconstruction, denoted by $\mathbf{Y}_u^e$. After combining the unlabeled samples with their corresponding predicted system responses, the autoencoder can be utilized for providing the latent variables, denoted by $\boldsymbol{\theta}_u^{'}$. It should be mentioned that the variables $\boldsymbol{\theta}_u^{'}$ are not the same as the actual latent variables $\boldsymbol{\theta}_u$ since the predicted system responses may not be accurate. However, $\boldsymbol{\theta}_u^{'}$ will definitely approach to the actual one if the estimated system responses from DFN and GP are close to the true responses. Under this circumstance, $\boldsymbol{\theta}_u^{'}$ tends to have similar values compared to the estimated $\boldsymbol{\theta}_u^e$, which has been already evaluated given only the unlabeled system input $\mathbf{X}_u$. In this work, we use the $nz$-dimension Euclidean distance to measure the divergence between $\boldsymbol{\theta}_u^{'}$ and $\boldsymbol{\theta}_u^e$ as follows

$$L(\boldsymbol{\theta}_u^{'}, \boldsymbol{\theta}_u^e) = \frac{1}{q}\sum_{i=1}^{q}\|\boldsymbol{\theta}_u^{'} - \boldsymbol{\theta}_u^e\| \quad (15)$$

The discrepancy $L(.,.)$ is considered as a measurement of the accuracy of system response prediction, which needs to be minimized during the semi-supervised learning process to enhance the accuracy of the system response prediction. For the labeled data, the mean square errors between the labels and the predictions also needs to be minimized during the training process of the DFN.

The training process of the proposed framework can be divided into three steps. Firstly, the autoencoder is trained based on the training data set $\mathbf{D}$, where the loss function is considered as the mean square error as shown in Eq. (6). With the available gradient information, the backpropagation algorithm is adopted while the "Adam" optimizer is used to train the autoencoder. In the second step, the latent variables $\boldsymbol{\theta}_t$ are first computed through the autoencoder, then the GP model is built given the training input $\boldsymbol{\theta}_t$ and training labels $\mathbf{Y}_t$ according to Eq. (7). It is worth noting that both the autoencoder and the GP model are fixed once they are trained. The reason lies in that 1) the fixed autoencoder can maintain the mapping relationship from the high-dimensional space to the low-dimensional latent space, and 2) the fixed GP captures the relationship between latent variables and system responses based on the labeled data. In the third step, the deep feedforward network is trained to minimize the aggregative loss, including both the prediction error and the discrepancy shown in Eq. (15), expressed as

$$\min_{\mathbf{w},\mathbf{b}} \alpha MSE(\boldsymbol{\theta}_t, \boldsymbol{\theta}_t^e) + \beta L(\boldsymbol{\theta}_u^{'}, \boldsymbol{\theta}_u^e) \quad (16)$$

where $\mathbf{w}$ and $\mathbf{b}$ represents the weights and biases matrix, respectively, $MSE(.,.)$ denotes the mean square error between the training labels and DFN predictions, $\alpha$ and $\beta$ are two coefficient pre-defined by the user. Note that the evaluation of the second term in Eq. (16) involves the computation of the autoencoder, the GP model, and the DFN, thus the gradient information may not be easily calculated.

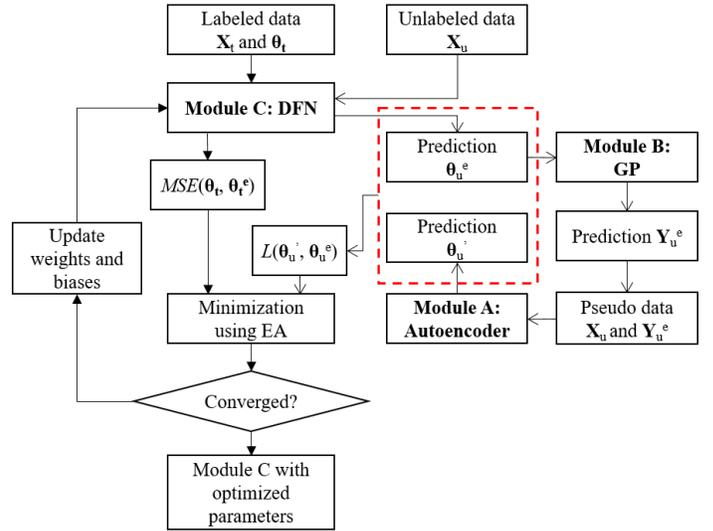

**FIGURE 2:** INFORMATION FLOW FOR THE PROPOSED SEMI-SUPERVISED LEARNING

Though gradient-based methods such as back-propagation have been widely used for deep neural networks, they may not applicable for problems without available gradient information. As an alternative option, evolutionary algorithms (EAs) have been utilized for such scenarios to train deep neural networks. Generally, evolutionary algorithms are known as probabilistic optimization techniques that are based on the principles of genetics and natural selection. When minimizing the loss function for deep neural networks, the differences between using evolutionary algorithms and gradient-based methods lie in that 1) EAs do not require any auxiliary information about the loss function, such as derivatives with respect to the weights, 2) EAs search for the optimal solution in a parallel way instead of searching from a single point. One of the benefits of employing the evolutionary algorithms is that they can significantly reduce the risk of falling into local minima. Therefore, EA is adopted for updating the parameters of the DFN. Starting from an initial population with random weights representing individuals, the aggregated loss value for each individual is computed based on Eq. (16). In each generation, the individuals with higher aggregated loss values are then replaced by the offspring of the individuals with lower aggregated loss values through crossover and mutation. The evolving process is iteratively until the convergence of the minimum aggregative loss value is converged, lead to a DFN with optimized parameters.

### 1.3 High Dimensional Reliability Analysis

Reliability analysis methods aim at evaluating the probability that an engineering system successfully performs its functionality with the consideration of various types of uncertainties. A limit state function $G(\mathbf{x})$ is generally used, where the system failure occurs when the limit state function value is less than zero. The probability of failure $P_f$ is then expressed as

$$P_f = \Pr[G(\mathbf{x}) < 0] = \int \cdots \int_{G(\mathbf{x})<0} f_x(\mathbf{x}) d\mathbf{x} \quad (17)$$

where $f_x(\mathbf{x})$ represents the joint probability density function of the input variables. Due to the multi-dimensional integral, directly using Eq. (17) for reliability analysis is usually prohibited and sampling methods such as Monte Carlo simulation (MCS) is commonly used as a substitution.

In the proposed approach, reliability for high-dimensional problems is approximated by the limit state function in the low-dimensional latent space, which is expressed as

$$P_f \approx \Pr[g(\boldsymbol{\theta}) < 0] = E\left[I_f(\boldsymbol{\theta})\right] \quad (18)$$

where $E[.]$ is the expectation operator, and $I_f(\boldsymbol{\theta}^e)$ is an indicator function to classify safe and failure samples. Assuming N random realizations of the original input variable x have been generated according to the input randomness, denoted as $\mathbf{X_m} = [\mathbf{x}_{m,1}, \mathbf{x}_{m,2}, \ldots, \mathbf{x}_{m,N}]$, the corresponding latent variables $\boldsymbol{\theta}^{em} = [\boldsymbol{\theta}_1^{em}, \boldsymbol{\theta}_2^{em}, \ldots, \boldsymbol{\theta}_N^{em}]$ can be approximated according to the optimized DFN. Then the system responses that correspond to $\mathbf{X_m}$ can be estimated by providing the estimated latent variables $\boldsymbol{\theta}^{em}$ to the constructed GP model, where the safe and failure samples can be classified by

$$I_f(\boldsymbol{\theta}_i^{em}) = \begin{cases} 1, & \mu_{gp}(\boldsymbol{\theta}_i^{em}) < 0 \quad (failure) \\ 0, & otherwise \quad (safe) \end{cases} \quad (19)$$

After predicting the responses for all the latent variables in $\boldsymbol{\theta}^{em}$, the system reliability can be easily computed as

$$R \approx \frac{\sum_{i=1}^{N} I_f(\boldsymbol{\theta}_i^{em})}{N} \quad (20)$$

### 2. CASE STUDIES

In this section, a high-dimensional mathematical example is used to test the performance of the proposed approach, where the limit state function is given as

$$G(\mathbf{x}) = 160.5 - \frac{(x_1^2 + 4)(x_2 - 1)}{20} + \cos(5x_1) - \sum_{i=1}^{20} x_i^2 \quad (21)$$

This example involves 20 random variables, which are assumed to be normally distributed with mean 2.86 and standard deviation 0.7. For data preparation, 150 samples are generated according to the input randomness, where the responses are directly evaluated based on Eq. (21), referred to as the labeled training data. In addition, 1000 random samples are generated as the unlabeled data, where the corresponding system responses remains unknown.

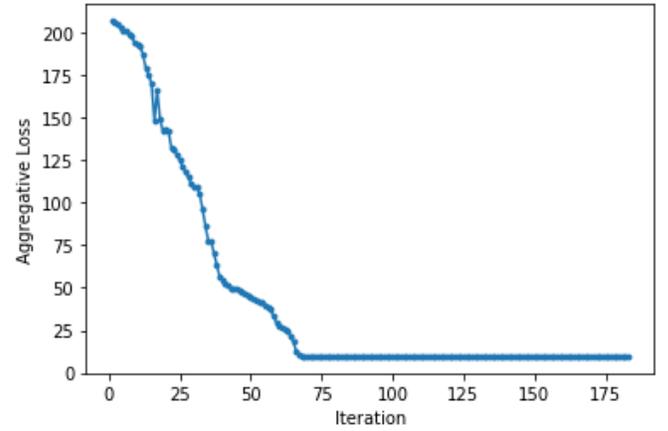

**FIGURE 3:** DFN UPDATING PROCESS USING EVOLUTIONARY ALGORITHM

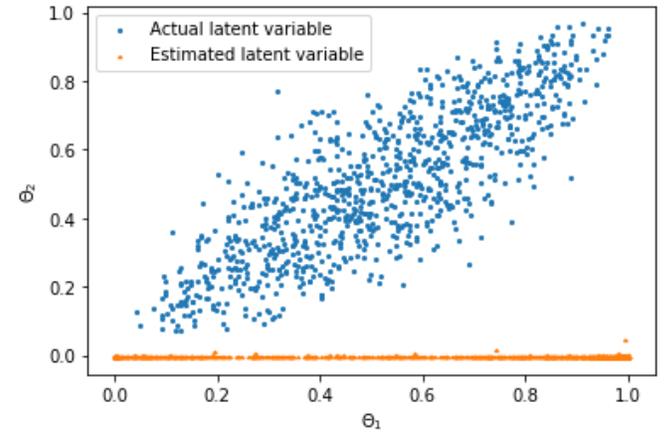

**FIGURE 4:** ESTIMATED AND ACTUAL LATENT VARIABLES FOR THE UNLABELED DATA BEFORE SEMI-SUPERVISED LEARNING

For the labeled data, the 150 inputs as well as the corresponding responses are fused into the training data set **D**, which is used to train the autoencoder. As a result, the actual latent variables that corresponding to the labeled data can be obtained. With available latent variables and their system

[Type here]

response, a GP model is established in the latent space for system response predictions. Following the procedure shown in subsection 1.2, the semi-supervised learning is performed based on both labeled and unlabeled data. The parameters of the DFN is optimized by employing the evolutionary algorithm. The semi-supervised learning stops at the 183th iteration, and the convergence of the aggregative loss function during the updating process is shown in Fig. (3). With the well-trained three critical modules, $10^5$ Monte Carlo simulation samples are utilized for reliability analysis, where the DFN is employed to estimate the latent variables and the system responses are predicted by the GP model. The reliability approximation by using the proposed approach is calculated as 0.7898, while the accurate reliability evaluated based on Eq. (21) is given as 0.7880. The result shows that the proposed approach can achieve an accurate reliability estimation with a relative error 0.2284%.

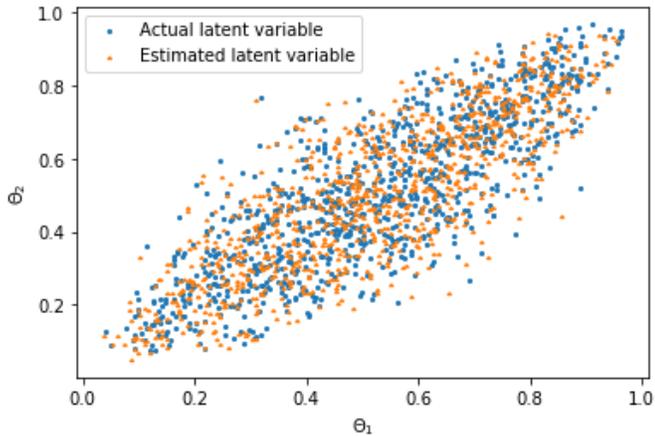

**FIGURE 5:** ESTIMATED AND ACTUAL LATENT VARIABLES FOR THE UNLABELED DATA AFTER SEMI-SUPERVISED LEARNING

To demonstrate the effectiveness of the proposed approach, the DFN prediction before semi-supervised learning is shown in Fig. (4). For illustration purpose, the actual latent variables of the unlabeled samples that are evaluated by providing both inputs and corresponding system responses into the autoencoder is depicted in Fig. (4) with circle markers, where the estimated latent variables from the DFN is represented by triangle marker. It clearly indicates that the initial DFN fails to reconstruct the latent space. After semi-supervised learning, the comparison of the estimated and the actual latent variables of the unlabeled data is shown in Fig. (5), where the accuracy has been significantly improved. For uncertainty quantification, the latent variables of the $10^5$ MCS samples are estimated by the DFN, then the responses are predicted by the GP model as shown in Fig. 6a), where safe and failure samples are classified based on the response prediction values. In Fig. 6b), the estimated latent variables are classified by using their actual system responses, which indicates that a failure surface exists in the latent space, reveals that transforming the high-dimensional problems into the low-dimensional latent space is reasonable for uncertainty propagation. The result show that the proposed approach can accurately capture the distinguishable limit state in the latent space, result in accurate reliability estimations.

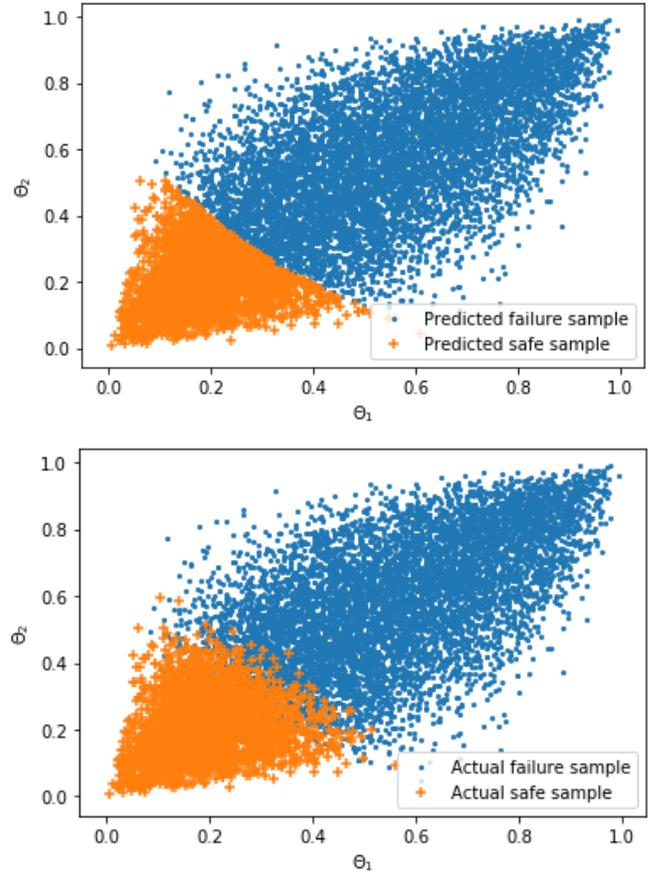

**FIGURE 6:** COMPARISON OF ESTIMATED LATENT VARIABLES WITH PREDICTED AND ACTUAL RESPONSES.

**CONCLUSION**
In this work, a novel semi-supervised learning method is proposed for uncertainty quantification of high-dimensional problems. With an autoencoder, a low-dimensional latent space is introduced for the purpose of dimension reduction, while a deep feedforward neural network and a GP model is utilized to transform the high-dimensional uncertainty quantification task into the latent space. To enhance the accuracy, labeled and unlabeled data are actively utilized during the semi-supervised learning process without incurring extra information of system responses. Instead of using back propagation, evolutionary algorithm is adopted for minimizing the aggregative loss. For uncertainty propagation, reliability analysis is then performed based on the combination of the three critical modules, where Monte Carlo simulation method is utilized for addressing the randomness of the system inputs. The results reveal that the proposed approach can be successfully applied for high-dimensional uncertainty quantification.